\documentclass[10pt]{article}
\usepackage[margin=1in]{geometry}
\usepackage[T1]{fontenc}
\usepackage{times}
\usepackage{amsmath,amssymb}
\usepackage{graphicx}
\usepackage{booktabs}
\usepackage[numbers,sort&compress]{natbib}
\usepackage[colorlinks=true,linkcolor=blue!50!black,citecolor=blue!50!black,urlcolor=blue!50!black]{hyperref}
\usepackage{xcolor}
\usepackage{caption}
\newcommand{\NOldNItems}{1264}
\newcommand{\NOldNLow}{156}

\newcommand{\NOldNLowBaseSixFour}{83}

\newcommand{\NOldLowRelEight}{0.89}

\newcommand{\NOldLowCeiling}{0.94}

\newcommand{\NOldLowCrossNoteacherROpdTeacher}{0.82}

\newcommand{\XStrataZeroZeroOneN}{156}

\newcommand{\XStrataZeroZeroOneLowCeiling}{0.94}

\newcommand{\XStrataZeroZeroOneLowBaseAcc}{0.14}

\newcommand{\XStrataZeroZeroOneLowaccCeiling}{0.96}

\newcommand{\XStrataZeroZeroOneLowaccBaseAcc}{0.57}

\newcommand{\XStrataZeroOneZeroFiveN}{192}

\newcommand{\XStrataZeroOneZeroFiveLowCeiling}{0.95}

\newcommand{\XStrataZeroOneZeroFiveLowBaseAcc}{0.11}

\newcommand{\XStrataZeroOneZeroFiveLowaccCeiling}{0.97}

\newcommand{\XStrataZeroOneZeroFiveLowaccBaseAcc}{0.64}

\newcommand{\XStrataZeroFiveOneN}{543}

\newcommand{\XStrataZeroFiveOneLowCeiling}{0.94}

\newcommand{\XStrataZeroFiveOneLowBaseAcc}{-0.79}

\newcommand{\XStrataZeroFiveOneLowaccCeiling}{0.97}

\newcommand{\XStrataZeroFiveOneLowaccBaseAcc}{0.90}

\newcommand{\XRelLowPerRunJustrlRelSingle}{0.77}

\newcommand{\XRelLowPerRunDeepscalerRelSingle}{0.52}

\newcommand{\XRelLowPerRunProrlRelSingle}{0.63}

\newcommand{\XRelLowPerRunStillThreeRelSingle}{0.34}

\newcommand{\XRelLowPerRunOpenRsThreeRelSingle}{0.16}

\newcommand{\XRelLowPerRunFastcurlRelSingle}{0.38}

\newcommand{\XRelLowPerRunOpdFullRelSingle}{0.81}

\newcommand{\XRelLowPerRunOpdOneShotRelSingle}{0.59}

\newcommand{\XRelLowNonlineageFiveCeilingFive}{0.89}

\newcommand{\XRelLowRankzCeiling}{0.94}

\newcommand{\XSimLowRho}{0.04}

\newcommand{\XSimLowKeptCeiling}{0.94}

\newcommand{\XRefFairLowWorstStepAvg}{0.44}

\newcommand{\XRefFairLowWorstStepAvgPartialK}{0.34}

\newcommand{\XRefFairLowMeanLikAvg}{0.26}

\newcommand{\XRefFairLowMeanLikAvgPartialK}{0.24}

\newcommand{\XRefFairLowRankOneSixAvg}{0.51}

\newcommand{\XRefFairLowRankOneSixAvgPartialK}{0.43}

\newcommand{\XRefFairLowLenAvg}{0.41}

\newcommand{\XRefFairLowLenAvgPartialK}{0.29}

\newcommand{\XSubsetLowN}{127}

\newcommand{\XSubsetLowCeiling}{0.95}

\newcommand{\XSubsetLowRidge}{0.49}

\newcommand{\XSubsetLowGbm}{0.42}

\newcommand{\XSubsetLowaccExplainedBestModelPoint}{0.54}

\newcommand{\XSubsetLowaccExplainedBestModelCiLo}{0.41}

\newcommand{\XSubsetLowaccExplainedBestModelCiHi}{0.68}

\newcommand{\XLoroLowMeanOthers}{0.67}

\newcommand{\XLoroLowByMOne}{0.54}

\newcommand{\XLoroLowByMTwo}{0.62}

\newcommand{\XLoroLowByMFour}{0.66}

\newcommand{\YVdLowAllFracProblem}{0.45}

\newcommand{\YVdLowNonlineageFracProblem}{0.28}

\newcommand{\YVdLowNonlineageFracNoisePost}{0.05}

\newcommand{\ZAgreeAllThreeTopic}{0.91}

\newcommand{\ZAgreeAllThreeInsight}{0.70}

\newcommand{\ZResidualTopicP}{0.93}

\newcommand{\ZModelWithoutLabelsExplainedBestModelPoint}{0.28}

\newcommand{\ZModelWithoutLabelsExplainedBestModelCiLo}{0.13}

\newcommand{\ZModelWithoutLabelsExplainedBestModelCiHi}{0.44}

\newcommand{\ZModelWithLabelsExplainedBestModelPoint}{0.27}

\newcommand{\ZModelWithLabelsExplainedBestModelCiLo}{0.14}

\newcommand{\ZModelWithLabelsExplainedBestModelCiHi}{0.45}

\newcommand{\ZModelLabelsOnlyExplainedBestModelPoint}{0.02}

\newcommand{\CNItems}{1532}

\newcommand{\CIndepN}{8}

\newcommand{\CIndepCeiling}{0.94}

\newcommand{\CVdIndepFracProblem}{0.87}

\newcommand{\CSubsetExplainedBestModelPoint}{0.41}

\newcommand{\CSubsetExplainedBestModelCiLo}{0.31}

\newcommand{\CSubsetExplainedBestModelCiHi}{0.53}

\newcommand{\CfNTf}{191}
\newcommand{\CfIndepCeiling}{0.90}

\newcommand{\CfVdIndepFracProblem}{0.36}

\newcommand{\CfVdIndepFracNoisePost}{0.07}

\newcommand{\CfSubsetCeiling}{0.91}

\newcommand{\CfSubsetRidge}{0.45}

\newcommand{\CfSubsetGbm}{0.38}

\newcommand{\CfSubsetAccCeiling}{0.96}

\newcommand{\CfSubsetAccExplainedBestModelPoint}{0.58}

\newcommand{\CfSubsetAccExplainedBestModelCiLo}{0.46}

\newcommand{\CfSubsetAccExplainedBestModelCiHi}{0.69}

\newcommand{\LtBaseOneN}{127}
\newcommand{\LtBaseOneCkpt}{0.52}

\newcommand{\LtBaseOneBaseOnly}{0.36}

\newcommand{\LtBaseOneRefSeven}{0.38}

\newcommand{\LtBaseOneDiffBaseOnly}{0.16}

\newcommand{\LtBaseOneDiffBaseOnlyCiLo}{0.06}

\newcommand{\LtBaseOneDiffBaseOnlyCiHi}{0.27}

\newcommand{\LtBaseOneDiffRefSeven}{0.13}

\newcommand{\LtBaseOneDiffRefSevenCiLo}{0.03}

\newcommand{\LtBaseOneDiffRefSevenCiHi}{0.23}

\newcommand{\LtBaseOneWinsBaseOnly}{8}
\newcommand{\LtBaseOneWinsRefSeven}{7}
\newcommand{\LtBaseOneNRuns}{8}

\newcommand{\LtBaseTwoAllDiffBaseOnly}{0.14}

\newcommand{\LtBaseTwoAllDiffBaseOnlyCiLo}{0.09}

\newcommand{\LtBaseTwoAllDiffBaseOnlyCiHi}{0.24}

\newcommand{\LtBaseTwoAllDiffRefSeven}{0.08}

\newcommand{\LtBaseTwoAllDiffRefSevenCiLo}{0.02}

\newcommand{\LtBaseTwoAllDiffRefSevenCiHi}{0.13}

\newcommand{\LtBaseTwoFmtN}{191}
\newcommand{\LtBaseTwoFmtCkpt}{0.34}

\newcommand{\LtBaseTwoFmtBaseOnly}{0.17}

\newcommand{\LtBaseTwoFmtRefSeven}{0.22}

\newcommand{\LtBaseTwoFmtDiffBaseOnly}{0.17}

\newcommand{\LtBaseTwoFmtDiffBaseOnlyCiLo}{0.07}

\newcommand{\LtBaseTwoFmtDiffBaseOnlyCiHi}{0.23}

\newcommand{\LtBaseTwoFmtDiffRefSeven}{0.12}

\newcommand{\LtBaseTwoFmtDiffRefSevenCiLo}{0.04}

\newcommand{\LtBaseTwoFmtDiffRefSevenCiHi}{0.21}

\newcommand{\LtBaseTwoFmtWinsBaseOnly}{9}
\newcommand{\LtBaseTwoFmtWinsRefSeven}{9}
\newcommand{\LtBaseTwoFmtNRuns}{10}

\newcommand{\PcBaseOneN}{68}
\newcommand{\PcBaseOneCeiling}{0.89}

\newcommand{\PcBaseOneShared}{0.28}

\newcommand{\PcBaseTwoN}{68}
\newcommand{\PcBaseTwoCeiling}{0.89}

\newcommand{\PcBaseTwoShared}{0.21}

\newcommand{\PcNPostcut}{268}

\newcommand{\FmRhoNoboxGain}{0.26}

\newcommand{\FmNLow}{395}
\newcommand{\FmNKept}{249}
\newcommand{\FmGainMin}{1.47}

\newcommand{\FmGainMax}{7.57}

\newcommand{\XfCeilingCrossfit}{0.84}

\newcommand{\XfSharedCrossfit}{0.35}

\newcommand{\XfLoroCkpt}{0.28}

\newcommand{\XfLoroBaseOnly}{0.16}

\newcommand{\XfLoroRefSeven}{0.21}

\newcommand{\AaCeilingAccIndep}{0.93}

\newcommand{\AaDiffRefSeven}{0.08}

\newcommand{\AbDiffRefSeven}{0.19}

\newcommand{\AbDiffRefSevenCiLo}{0.12}

\newcommand{\AbDiffRefSevenCiHi}{0.29}

\newcommand{\ExAGainExplained}{0.30}

\newcommand{\ExAGainExplainedCiLo}{0.19}

\newcommand{\ExAGainExplainedCiHi}{0.48}

\newcommand{\ExBGainExplained}{0.24}

\newcommand{\ExBGainExplainedCiLo}{0.14}

\newcommand{\ExBGainExplainedCiHi}{0.38}

\newcommand{\UABaseAccRaw}{0.19}

\newcommand{\UABaseAccLo}{0.01}

\newcommand{\UABaseAccHi}{0.35}

\newcommand{\UAMeanLikRaw}{0.26}

\newcommand{\UAMeanLikLo}{0.08}

\newcommand{\UAMeanLikHi}{0.44}

\newcommand{\UASumLpRaw}{0.43}

\newcommand{\UASumLpLo}{0.26}

\newcommand{\UASumLpHi}{0.59}

\newcommand{\UAWorstStepRaw}{0.47}

\newcommand{\UAWorstStepLo}{0.31}

\newcommand{\UAWorstStepHi}{0.61}

\newcommand{\UARankOneSixRaw}{-0.51}

\newcommand{\UARankOneSixLo}{-0.64}

\newcommand{\UARankOneSixHi}{-0.35}

\newcommand{\UAShortSolRaw}{-0.36}

\newcommand{\UAShortSolLo}{-0.52}

\newcommand{\UAShortSolHi}{-0.18}

\newcommand{\UARefSevenAccRaw}{0.52}

\newcommand{\UARefSevenAccLo}{0.36}

\newcommand{\UARefSevenAccHi}{0.65}

\newcommand{\UAN}{127}
\newcommand{\UBBaseAccRaw}{0.29}

\newcommand{\UBBaseAccLo}{0.16}

\newcommand{\UBBaseAccHi}{0.43}

\newcommand{\UBMeanLikRaw}{0.18}

\newcommand{\UBMeanLikLo}{0.05}

\newcommand{\UBMeanLikHi}{0.32}

\newcommand{\UBSumLpRaw}{0.32}

\newcommand{\UBSumLpLo}{0.17}

\newcommand{\UBSumLpHi}{0.46}

\newcommand{\UBWorstStepRaw}{0.21}

\newcommand{\UBWorstStepLo}{0.07}

\newcommand{\UBWorstStepHi}{0.33}

\newcommand{\UBRankOneSixRaw}{-0.10}

\newcommand{\UBRankOneSixLo}{-0.24}

\newcommand{\UBRankOneSixHi}{0.05}

\newcommand{\UBShortSolRaw}{-0.13}

\newcommand{\UBShortSolLo}{-0.27}

\newcommand{\UBShortSolHi}{0.01}

\newcommand{\UBRefSevenAccRaw}{0.45}

\newcommand{\UBRefSevenAccLo}{0.32}

\newcommand{\UBRefSevenAccHi}{0.57}

\newcommand{\UBN}{191}
\newcommand{\UBAllCeilingFmt}{0.91}

\newcommand{\LoALofoMin}{0.83}

\newcommand{\LoALofoMax}{0.90}

\newcommand{\LoBLofoMin}{0.88}

\newcommand{\LoBLofoMax}{0.91}

\newcommand{\XbN}{152}
\newcommand{\XbCrossR}{0.25}

\newcommand{\XbCrossCiLo}{0.09}

\newcommand{\XbCrossCiHi}{0.42}

\newcommand{\XbCeilA}{0.90}

\newcommand{\XbCeilB}{0.91}

\newcommand{\XaCeilingCrossfit}{0.80}

\newcommand{\XaSharedCrossfit}{0.24}

\newcommand{\XaLoroCkpt}{0.37}

\newcommand{\XaLoroBaseOnly}{0.28}

\newcommand{\XaLoroRefSeven}{0.28}

\newcommand{\PeAPearsonCeiling}{0.89}

\newcommand{\PeBPearsonCeiling}{0.92}

\newcommand{\CoACombo}{0.33}

\newcommand{\CoAWinsCombo}{8}
\newcommand{\CoANRuns}{8}

\newcommand{\CoAPairMax}{0.74}

\newcommand{\CoAPairMedian}{0.50}

\newcommand{\CoBCombo}{0.17}

\newcommand{\CoBWinsCombo}{9}
\newcommand{\CoBNRuns}{10}

\newcommand{\CoBPairMax}{0.68}

\newcommand{\CoBPairMedian}{0.33}

\title{\textbf{What Will Post-Training Fix?}\\[6pt]
\large Per-Problem Gains Are Shared Across Independent RL Runs,\\ and Existing Checkpoints Predict Them Better Than A Priori Signals}
\author{%
Xiaoxian Duan$^{1,2,3}$\\[3pt]
$^1$Institute of Automation, Chinese Academy of Sciences, Beijing, China\\
$^2$Zhongguancun Academy, Beijing, China\\
$^3$Z.ai\\
\texttt{duanxiaoxian2026@ia.ac.cn}
}
\date{}

\begin{document}
\maketitle

\begin{abstract}
Data selection, curricula and the evaluation of post-training recipes all assume that we can tell, before training, which problems a model will improve on. We test this assumption directly. For two base models, DeepSeek-R1-Distill-Qwen-1.5B and Qwen2.5-Math-1.5B, we evaluate eighteen post-training runs on up to \CNItems{} competition math problems with many samples per problem, and compare signals available before training against a \emph{noise ceiling} derived from the agreement between disjoint subsets of runs. Three findings hold on both base models. \textbf{What post-training fixes is shared:} independent runs agree on which rarely solved problems improve, with a noise ceiling of about $0.9$, yet the two base models agree with each other only at $\rho=\XbCrossR$---the shared component belongs to the base model, not to the problem. \textbf{A priori signals capture a minority of it:} base pass rate, the likelihood of a correct solution, a larger model's pass rate and their combinations explain only $\ExAGainExplained$ and $\ExBGainExplained$ of the explainable variance in gains. \textbf{Existing checkpoints are the better predictor:} the per-problem gains of a single checkpoint from another family predict a new run better than every a priori signal, alone or combined ($\LtBaseOneCkpt$ vs.\ $\CoACombo$ and $\LtBaseTwoFmtCkpt$ vs.\ $\CoBCombo$ against the combined signals). The conclusions hold on problems from 2025--2026 competitions and when the baselines on the two sides of every comparison are estimated independently. We propose the noise ceiling as a standard companion to per-problem signals.
\end{abstract}

\section{Introduction}

Post-training with verifiable rewards has become the main lever for improving reasoning in language models~\citep{deepseek2025r1,shao2024deepseekmath,yu2025dapo}, and with it a set of per-problem decisions has become routine. Pipelines discard problems whose rollouts are all wrong; curricula order problems by the model's current pass rate; recent methods route on pass rate and then keep the verified solutions the model finds most likely~\citep{xu2026trek}; and practitioners compare recipes by which held-out problems they improve. Each of these decisions presumes an answer to a basic question: \emph{before post-training, can we tell which problems a model will improve on?}

A common intuition is a notion we call \emph{closeness}. A model is expected to gain on problems it is already near to solving, as measured by its pass rate or by the likelihood it assigns to a correct solution. Closeness is intuitive and cheap, but it has rarely been evaluated against a yardstick that separates a weak signal from an unpredictable target. Whether a run improves a problem depends on the run's data, algorithm and seed and on finite evaluation, so a weak predictor and an unpredictable target look the same. And most studies observe one run, so what is shared across recipes cannot be separated from what is specific to one.

We resolve both difficulties with a simple design and evaluate closeness against the right yardstick. Holding the base model fixed, we collect many post-training runs released by different groups and evaluate each of them, with many samples per problem, on the same held-out problems. Agreement between disjoint subsets of runs gives a \emph{noise ceiling}: the highest correlation any signal can be expected to reach with the runs' consensus. A signal can then be judged by the fraction of the explainable variance it captures, rather than by a raw correlation of unknown meaning. We carry out this design on two base models with different post-training histories, DeepSeek-R1-Distill-Qwen-1.5B (eight runs) and Qwen2.5-Math-1.5B (ten runs), and report only conclusions that hold on both.

\begin{figure}[t]
\centering
\includegraphics[width=\linewidth]{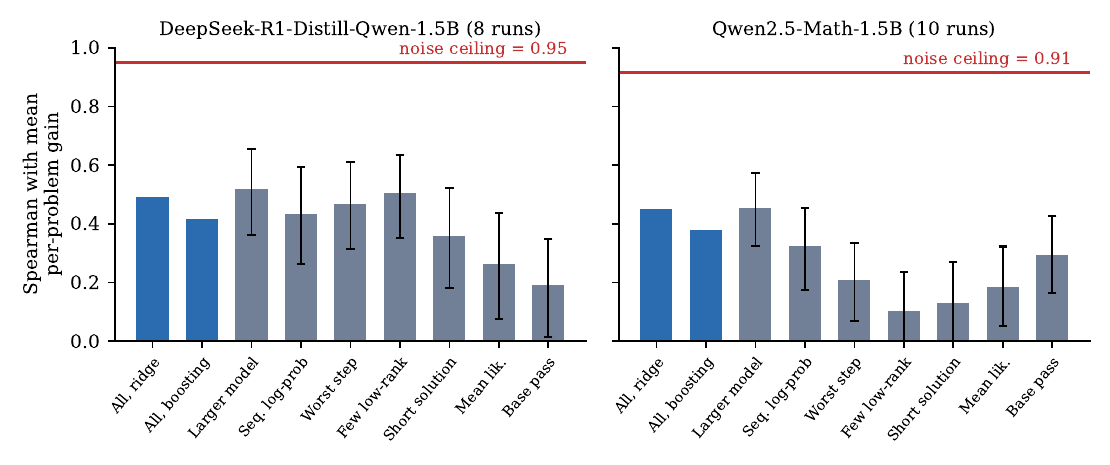}
\caption{\textbf{A priori signals reach a minority of what is predictable, on both base models.} All bars use the \UAN{} (Base A) and \UBN{} (Base B) rarely solved problems with a reference solution. ``Few low-rank'': fewer reference-solution steps outside the base model's top-16. Bars: Spearman correlation between each a priori signal and the mean per-problem gain of the runs on problems the base model rarely solves, with 95\% bootstrap intervals; blue bars combine all signals with ten-fold cross-validation. Red line: noise ceiling on the same problems. Each signal is named in the direction associated with larger gains.}
\label{fig:ceiling}
\end{figure}

\paragraph{Findings.}
\begin{enumerate}
\item \textbf{What post-training fixes is shared across independent runs.} Runs from different groups and with different algorithms agree on which rarely solved problems improve. The noise ceiling for their consensus is about $0.9$ on both base models, the agreement holds on recent competition problems, and it is specific to the base model: gains on the two bases correlate at only $\XbCrossR$ (Section~\ref{sec:shared}).
\item \textbf{A priori signals capture a minority of the shared component.} The strongest single signal on each base reaches about half the ceiling, and the best single signal or cross-validated combination explains $\ExAGainExplained$ and $\ExBGainExplained$ of the explainable variance in gains (Figure~\ref{fig:ceiling}, Section~\ref{sec:signals}). The missing part lives in the base model's behaviour on a problem, not in what the problem looks like.
\item \textbf{Existing checkpoints are the better predictor.} A single checkpoint from another family predicts which problems a new run will improve better than the best base-model signal for 17 of 18 held-out runs, and better than a larger model's pass rate for 16 of 18 (Section~\ref{sec:loro}).
\end{enumerate}
Together these results change how per-problem prediction in post-training should be thought of. What a run will fix is not noise: it is a stable property of the base model and the problem, reproduced by runs from different groups and algorithms and distinct from one base model to the next. Yet the closeness statistics on which data selection and curricula are built see only a minority of it, while the outcomes of runs that already exist see much more. The gap between the two is now measurable, and closing it is a concrete target for the next generation of selection methods.

\section{Setup}
\label{sec:setup}

\paragraph{Base models and runs.} \emph{Base A} is DeepSeek-R1-Distill-Qwen-1.5B~\citep{deepseek2025r1}. Its eight runs are six public RLVR checkpoints from six groups---JustRL~\citep{he2025justrl}, DeepScaleR~\citep{luo2025deepscaler}, ProRL~\citep{liu2025prorl}, STILL-3~\citep{min2024still}, Open-RS3~\citep{dang2025openrs} and FastCuRL~\citep{song2025fastcurl}---and two on-policy distillation (OPD) runs that we train from JustRL following~\citet{fu2026oneshot}. Our full-data OPD run reproduces the reported result (69.5 vs.\ 69.8 macro accuracy on MATH-500/AMC23/AIME25). \emph{Base B} is Qwen2.5-Math-1.5B~\citep{yang2024qwen25math}. Its ten runs are public RL checkpoints from eight groups, including Oat-Zero~\citep{liu2025understanding}, three One-Shot RLVR checkpoints~\citep{wang2025oneshot}, ExGRPO~\citep{zhan2025exgrpo}, ReLIFT~\citep{ma2025relift} and a spurious-rewards checkpoint~\citep{shao2025spurious}; the full list is in Appendix~\ref{app:details}. We call runs that share a lineage a \emph{family}: JustRL and the two OPD runs distilled from it on Base A, and the three One-Shot checkpoints on Base B. \emph{Independent runs} come from different families: on Base B we keep one checkpoint per family (eight runs); on Base A we drop the JustRL family entirely, because our OPD runs inherit their teacher's behaviour (five runs).

\paragraph{Problems.} We pool MATH-500~\citep{hendrycks2021math}, AMC23, AIME24, AIME25, HMMT25, OlympiadBench~\citep{he2024olympiadbench} and a subset of Omni-MATH~\citep{gao2024omnimath}, and remove every problem whose normalised statement matches a problem in DAPO-17k or in five public RL training sets used by the Base A runs, leaving \NOldNItems{} problems. We add \PcNPostcut{} problems from 2025--2026 competitions collected by MathArena~\citep{balunovic2025matharena}, which are newer than the training data of the Base A checkpoints; we analyse them separately, as a held-out check, and for each base model use those it rarely solves; Base B is evaluated on all \CNItems{} problems.

\paragraph{Sampling and outcomes.} Base A models generate up to 31{,}744 tokens and Base B models up to 3{,}072 tokens (the context limit of Qwen2.5-Math), with temperature 0.7 and top-$p$ 0.9; all outputs are graded by the same rule-based verifier. Each base model is sampled 256 times per problem (64 times for one batch of Base A problems, Appendix~\ref{app:details}) and each post-trained model 64 times. We study problems the base model rarely solves, with base pass rate in $(0, 25/256]$, and take as outcome the gain in pass rate from base to post-trained model. On Base B, part of the gain on many problems comes from the base model failing to finish with a final boxed answer, which RL training corrects: the fraction of base samples without a final answer correlates with gain at $\rho=\FmRhoNoboxGain$. Because this reflects answer formatting rather than the problem, our main analysis on Base B uses the \FmNKept{} problems on which at most a quarter of base samples lack a final answer; results on all \FmNLow{} problems are reported in Appendix~\ref{app:robust}. The main analysis covers \NOldNLow{} problems on Base A and \FmNKept{} on Base B.

\paragraph{A priori signals.} \emph{Base pass rate} comes from the base samples. Likelihood statistics need a correct solution that the base model did not write: we sample a larger model from the same family (DeepSeek-R1-Distill-Qwen-7B for Base A, Qwen2.5-Math-7B-Instruct for Base B) 64 times per problem, keep up to four correct solutions, and teacher-force the base model on them. From token log-probabilities $\ell_t$ and ranks $r_t$ we compute the \emph{mean likelihood} (the mean of $\ell_t$ over the central 90\% of tokens, a symmetric variant of the trimmed statistic~\citet{xu2026trek} use to rank verified solutions), the \emph{sequence log-probability} $\sum_t \ell_t$, the \emph{worst-step log-probability} $\min_t \ell_t$, the number of steps with $r_t>16$, and the solution length. The \emph{larger-model pass rate} is the 7B model's own accuracy. Likelihood statistics are available on the \XSubsetLowN{} (Base A) and \CfNTf{} (Base B) problems for which the larger model found a correct solution.

\section{A noise ceiling for per-problem prediction}
\label{sec:ceiling}

Let $y_{p,j}$ be the gain of run $j$ on problem $p$ and $\bar y_p$ its mean over a set of runs. For every split of the runs into two equal halves we compute the Spearman correlation between the half means across problems and average, obtaining the split-half agreement $r$; with an odd number of runs we leave one run out of each split. The Spearman--Brown formula~\citep{spearman1904proof,brown1910some} extends $r$ to the reliability $R$ of the mean over all runs; intuitively, it answers how reliable the average of all runs is, given how well two halves of them agree. Under the classical model in which each run's outcome is a shared component plus run-specific deviation, no a priori predictor can be expected to correlate with $\bar y_p$ more than $\sqrt{R}$, which we call the noise ceiling; the construction is standard in neuroscience~\citep{nili2014toolbox} and has been used to bound per-item predictability for language models~\citep{zhang2026illusion}. A signal with correlation $\rho$ captures a fraction $\rho^2/R$ of the explainable variance. Pearson correlations give the same ceilings ($\PeAPearsonCeiling$ and $\PeBPearsonCeiling$). We complement the ceiling with a variance decomposition of single-run gains, in which the shared problem effect is the average covariance between different runs and sampling noise is the binomial variance of a 64-sample estimate. All intervals are 95\% bootstrap intervals over problems.

\paragraph{Independent baselines.} Every run's gain subtracts the same estimate of the base pass rate. To make the comparison between runs fully independent, we also estimate it separately for the two sides of each comparison. On Base B we split the 256 base samples into 128 to select problems and compute base-model signals and two sets of 64 that serve as the two baselines, averaging over ten random splits; on Base A we draw 64 additional base samples per problem with a new seed and use them as the baseline for one side. We also report post-training accuracy as an outcome, which subtracts nothing.

\section{What post-training fixes is shared across independent runs}
\label{sec:shared}

\begin{table}[t]
\centering
\small
\begin{tabular}{lcc}
\toprule
& Base A: R1-Distill-1.5B & Base B: Qwen2.5-Math-1.5B \\
\midrule
Rarely solved problems & \NOldNLow & \FmNKept \\
Independent runs & 5 & \CIndepN \\
Noise ceiling, independent runs & $\XRelLowNonlineageFiveCeilingFive$ & $\CfIndepCeiling$ \\
Noise ceiling, all runs & $\NOldLowCeiling$ & $\UBAllCeilingFmt$ \\
Shared problem effect (fraction of single-run variance) & $\YVdLowNonlineageFracProblem$ & $\CfVdIndepFracProblem$ \\
Sampling noise (fraction of single-run variance) & $\YVdLowNonlineageFracNoisePost$ & $\CfVdIndepFracNoisePost$ \\
\midrule
2025--2026 competition problems only: $N$ & \PcBaseOneN & \PcBaseTwoN \\
\quad noise ceiling, independent runs & $\PcBaseOneCeiling$ & $\PcBaseTwoCeiling$ \\
\quad shared problem effect & $\PcBaseOneShared$ & $\PcBaseTwoShared$ \\
\midrule
Leave one family out: range of noise ceiling & $\LoALofoMin$--$\LoALofoMax$ & $\LoBLofoMin$--$\LoBLofoMax$ \\
\midrule
Independent baselines: noise ceiling & $\XaCeilingCrossfit$ & $\XfCeilingCrossfit$ \\
\quad shared problem effect & $\XaSharedCrossfit$ & $\XfSharedCrossfit$ \\
Post-training accuracy as outcome: noise ceiling & $\AaCeilingAccIndep$ & $\CfSubsetAccCeiling$ \\
\bottomrule
\end{tabular}
\caption{\textbf{Independent runs agree on which problems improve.} Noise ceilings and variance decomposition of per-problem gains on problems the base model rarely solves. Independent runs as defined in Section~\ref{sec:setup}. The last block uses independent baselines (Section~\ref{sec:ceiling}).}
\label{tab:shared}
\end{table}

Table~\ref{tab:shared} summarises the central measurement, and the message is the same on both base models: runs from different groups agree on which rarely solved problems improve. The mean gain over independent runs has a noise ceiling of $\XRelLowNonlineageFiveCeilingFive$ on Base A and $\CfIndepCeiling$ on Base B. Any single run is a noisy reading---most of its per-problem variation is specific to that run---but a substantial shared problem effect remains: $\YVdLowNonlineageFracProblem$ and $\CfVdIndepFracProblem$ of a single run's variance, against $\YVdLowNonlineageFracNoisePost$ and $\CfVdIndepFracNoisePost$ for sampling noise.

The agreement is robust. Leaving each family out in turn keeps the ceiling between $\LoALofoMin$ and $\LoALofoMax$ on Base A and between $\LoBLofoMin$ and $\LoBLofoMax$ on Base B. With independent baselines for the two halves the ceilings are $\XaCeilingCrossfit$ and $\XfCeilingCrossfit$. And it holds on 2025--2026 competition problems, where the ceiling is $\PcBaseOneCeiling$ on both bases. The two base models differ in their post-training history---one is distilled from long reasoning traces, the other is a math-specialised pretrained model on which RL elicits reasoning from scratch---yet the same structure appears in both.

\paragraph{The shared component belongs to the base model.} Both bases were evaluated on the same original pool, so we can ask whether the problems one base model improves on are the problems the other improves on. On the \XbN{} problems that both bases solve at a rate in $(0, 0.5]$, the runs of each base agree strongly among themselves (ceilings $\XbCeilA$ and $\XbCeilB$ on exactly these problems), yet the mean gain on Base A and the mean gain on Base B correlate at only $\rho=\XbCrossR$ [$\XbCrossCiLo$, $\XbCrossCiHi$]. What post-training fixes is a property of the base model's relation to the problem: it is reproducible within a base model and largely distinct across base models.

\paragraph{It is independent of training-data overlap and of problem type.} On Base A, the maximum character n-gram TF-IDF similarity between a problem and the union of public RL training sets is unrelated to its gain ($\rho=\XSimLowRho$), and after removing the quarter of problems most similar to any training problem, the remaining problems retain an eight-run ceiling of $\XSimLowKeptCeiling$. An LLM annotated each problem three times, blind to outcomes, for topic, answer type, number of solution steps, case analysis, computational load, need for a key insight, need for a construction and answer-format ambiguity (three-way agreement $\ZAgreeAllThreeInsight$--$\ZAgreeAllThreeTopic$); a model trained on all labels captures $\ZModelLabelsOnlyExplainedBestModelPoint$ of the explainable variance. The shared component is carried by the base model's behaviour on the problem, not by what the problem looks like.

\section{A priori signals capture a minority of the shared component}
\label{sec:signals}

\begin{table}[t]
\centering
\small
\begin{tabular}{lcc}
\toprule
A priori signal & Base A: R1-Distill-1.5B & Base B: Qwen2.5-Math-1.5B \\
\midrule
Base pass rate & $\UABaseAccRaw$ $[\UABaseAccLo, \UABaseAccHi]$ & $\UBBaseAccRaw$ $[\UBBaseAccLo, \UBBaseAccHi]$ \\
Mean likelihood (trimmed) & $\UAMeanLikRaw$ $[\UAMeanLikLo, \UAMeanLikHi]$ & $\UBMeanLikRaw$ $[\UBMeanLikLo, \UBMeanLikHi]$ \\
Sequence log-probability & $\UASumLpRaw$ $[\UASumLpLo, \UASumLpHi]$ & $\UBSumLpRaw$ $[\UBSumLpLo, \UBSumLpHi]$ \\
Worst-step log-probability & $\UAWorstStepRaw$ $[\UAWorstStepLo, \UAWorstStepHi]$ & $\UBWorstStepRaw$ $[\UBWorstStepLo, \UBWorstStepHi]$ \\
Steps outside base top-16 & $\UARankOneSixRaw$ $[\UARankOneSixLo, \UARankOneSixHi]$ & $\UBRankOneSixRaw$ $[\UBRankOneSixLo, \UBRankOneSixHi]$ \\
Solution length & $\UAShortSolRaw$ $[\UAShortSolLo, \UAShortSolHi]$ & $\UBShortSolRaw$ $[\UBShortSolLo, \UBShortSolHi]$ \\
Larger-model pass rate & $\UARefSevenAccRaw$ $[\UARefSevenAccLo, \UARefSevenAccHi]$ & $\UBRefSevenAccRaw$ $[\UBRefSevenAccLo, \UBRefSevenAccHi]$ \\
\midrule
All signals, ridge (CV) & $\XSubsetLowRidge$ & $\CfSubsetRidge$ \\
All signals, boosting (CV) & $\XSubsetLowGbm$ & $\CfSubsetGbm$ \\
Noise ceiling on the same problems & $\XSubsetLowCeiling$ & $\CfSubsetCeiling$ \\
\midrule
Explainable variance captured, gain & $\ExAGainExplained$ $[\ExAGainExplainedCiLo, \ExAGainExplainedCiHi]$ & $\ExBGainExplained$ $[\ExBGainExplainedCiLo, \ExBGainExplainedCiHi]$ \\
Explainable variance captured, post-training acc. & $\XSubsetLowaccExplainedBestModelPoint$ $[\XSubsetLowaccExplainedBestModelCiLo, \XSubsetLowaccExplainedBestModelCiHi]$ & $\CfSubsetAccExplainedBestModelPoint$ $[\CfSubsetAccExplainedBestModelCiLo, \CfSubsetAccExplainedBestModelCiHi]$ \\
\bottomrule
\end{tabular}
\caption{\textbf{A priori signals against the noise ceiling.} Spearman correlation of each signal with the mean per-problem gain over all runs, on the \UAN{} (Base A) and \UBN{} (Base B) rarely solved problems for which the larger model found a correct solution; brackets are 95\% bootstrap intervals. Signs are those of the raw signal (longer solutions and more low-rank steps go with smaller gains). ``Explainable variance captured'' is $\rho^2/R$ for the best single signal or cross-validated combination, whichever is higher.}
\label{tab:signals}
\end{table}

Table~\ref{tab:signals} and Figure~\ref{fig:ceiling} compare a priori signals with the ceiling. The pattern is the same on both base models. The closeness signals in widest use are weak. Base pass rate reaches $\rho=\UABaseAccRaw$ and $\UBBaseAccRaw$, and a trimmed mean likelihood of the kind TREK uses to rank verified solutions reaches $\UAMeanLikRaw$ and $\UBMeanLikRaw$, below both the sequence log-probability and the larger model's pass rate. Among the strongest signals is the larger model's pass rate ($\UARefSevenAccRaw$ and $\UBRefSevenAccRaw$), which is not a closeness measure of the base model at all but a measure of whether the problem is solvable within the model family. Combining every signal with cross-validated ridge regression or gradient boosting adds nothing beyond the best single signal: the signals see the same part of the shared component. Taking the better of the best single signal and the best combination, a priori signals capture $\ExAGainExplained$ [$\ExAGainExplainedCiLo$, $\ExAGainExplainedCiHi$] and $\ExBGainExplained$ [$\ExBGainExplainedCiLo$, $\ExBGainExplainedCiHi$] of the explainable variance in gains.

\paragraph{Gain versus accuracy.} Gain is the outcome that matters for deciding where training helps, but signals are often evaluated against post-training accuracy instead. Accuracy is easier to predict, because it inherits the base model's ability: the signals capture $\XSubsetLowaccExplainedBestModelPoint$ and $\CfSubsetAccExplainedBestModelPoint$ of the explainable variance in accuracy, roughly twice the fraction for gains. Predicting where a model will end up is thus considerably easier than predicting how far it will move, and conclusions drawn from accuracy overstate how well a signal anticipates the effect of training.

\section{Existing checkpoints predict a new run better than a priori signals}
\label{sec:loro}

\begin{figure}[t]
\centering
\begin{minipage}{0.48\linewidth}\centering
\includegraphics[width=\linewidth]{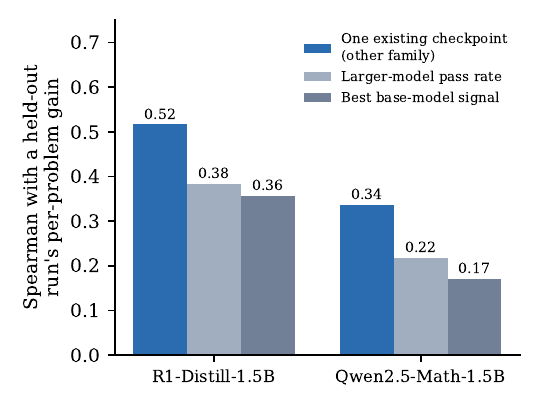}
\end{minipage}\hfill
\begin{minipage}{0.49\linewidth}\centering
\small
\begin{tabular}{lcc}
\toprule
& Base A & Base B \\
\midrule
One checkpoint, other family & $\LtBaseOneCkpt$ & $\LtBaseTwoFmtCkpt$ \\
Best base-model signal & $\LtBaseOneBaseOnly$ & $\LtBaseTwoFmtBaseOnly$ \\
\quad difference & $+\LtBaseOneDiffBaseOnly$ & $+\LtBaseTwoFmtDiffBaseOnly$ \\
\quad 95\% interval & $[\LtBaseOneDiffBaseOnlyCiLo, \LtBaseOneDiffBaseOnlyCiHi]$ & $[\LtBaseTwoFmtDiffBaseOnlyCiLo, \LtBaseTwoFmtDiffBaseOnlyCiHi]$ \\
\quad held-out runs won & \LtBaseOneWinsBaseOnly/\LtBaseOneNRuns & \LtBaseTwoFmtWinsBaseOnly/\LtBaseTwoFmtNRuns \\
Larger-model pass rate & $\LtBaseOneRefSeven$ & $\LtBaseTwoFmtRefSeven$ \\
\quad difference & $+\LtBaseOneDiffRefSeven$ & $+\LtBaseTwoFmtDiffRefSeven$ \\
\quad 95\% interval & $[\LtBaseOneDiffRefSevenCiLo, \LtBaseOneDiffRefSevenCiHi]$ & $[\LtBaseTwoFmtDiffRefSevenCiLo, \LtBaseTwoFmtDiffRefSevenCiHi]$ \\
\quad held-out runs won & \LtBaseOneWinsRefSeven/\LtBaseOneNRuns & \LtBaseTwoFmtWinsRefSeven/\LtBaseTwoFmtNRuns \\
All a priori signals, CV ridge & $\CoACombo$ & $\CoBCombo$ \\
\quad held-out runs won & \CoAWinsCombo/\CoANRuns & \CoBWinsCombo/\CoBNRuns \\
\bottomrule
\end{tabular}
\end{minipage}
\caption{\textbf{A single existing checkpoint is a better guide than any a priori signal, alone or combined.} Each run is held out in turn; we correlate its per-problem gains with those of one checkpoint from another family (averaged over such checkpoints), with the best base-model signal, and with the larger model's pass rate. The base-model signal is selected on the mean of the remaining runs, so the held-out run never influences the choice. Problems with a reference solution: \LtBaseOneN{} on Base A, \LtBaseTwoFmtN{} on Base B.}
\label{fig:loro}
\end{figure}

If what post-training fixes is shared across runs, then the runs that already exist should tell us what a new run will fix. We test this directly. Each run is held out in turn, and we ask how well its per-problem gains are predicted by three sources: a single checkpoint from another family, the best signal computable from the base model, and the larger model's pass rate (Figure~\ref{fig:loro}). To keep the comparison fair, the base-model signal is chosen on the remaining runs rather than on the held-out one.

On both base models a single checkpoint from another family is the best predictor. It exceeds the best base-model signal by $\LtBaseOneDiffBaseOnly$ [$\LtBaseOneDiffBaseOnlyCiLo$, $\LtBaseOneDiffBaseOnlyCiHi$] on Base A and $\LtBaseTwoFmtDiffBaseOnly$ [$\LtBaseTwoFmtDiffBaseOnlyCiLo$, $\LtBaseTwoFmtDiffBaseOnlyCiHi$] on Base B, and wins for $\LtBaseOneWinsBaseOnly$ of $\LtBaseOneNRuns$ and $\LtBaseTwoFmtWinsBaseOnly$ of $\LtBaseTwoFmtNRuns$ held-out runs. It also beats the larger model's pass rate by $\LtBaseOneDiffRefSeven$ and $\LtBaseTwoFmtDiffRefSeven$, with intervals excluding zero, and a ridge model that combines all a priori signals and is fitted with ten-fold cross-validation to each held-out run's own gains ($\CoACombo$ and $\CoBCombo$; the checkpoint wins for \CoAWinsCombo{} of \CoANRuns{} and \CoBWinsCombo{} of \CoBNRuns{} held-out runs). Across individual pairs of checkpoints from different families the median correlation is $\CoAPairMedian$ and $\CoBPairMedian$, and the best-matched pairs reach $\CoAPairMax$ and $\CoBPairMax$. On Base A, prediction improves further as more checkpoints are pooled (Appendix~\ref{app:robust}; that analysis draws runs from any family). This result is the predictive face of the shared component of Section~\ref{sec:shared}: a single run's gains contain the shared component plus run-specific variation, so they act as an achievable lower bound on the ceiling that requires no model of what the shared component is.

\paragraph{Independent baselines.} With disjoint base samples for the checkpoint and the held-out run, the checkpoint reaches $\XaLoroCkpt$ on Base A, against $\XaLoroBaseOnly$ for the best base-model signal and $\XaLoroRefSeven$ for the larger model, and $\XfLoroCkpt$ on Base B, against $\XfLoroBaseOnly$ and $\XfLoroRefSeven$. With post-training accuracy as outcome, a checkpoint from another family exceeds the larger model's pass rate by $\AaDiffRefSeven$ on Base A and $\AbDiffRefSeven$ [$\AbDiffRefSevenCiLo$, $\AbDiffRefSevenCiHi$] on Base B.

\paragraph{A practical recipe.} To anticipate which held-out problems a new post-training run of a base model will improve, evaluate the base model and one existing checkpoint of it with a few dozen samples per problem, and rank problems by the checkpoint's gain. No reference solutions, larger models or likelihood computations are needed, and the ranking is more informative than any a priori statistic we tested. For base models that are being post-trained for the first time, the noise ceiling shows how much room is left for a priori signals to grow into.

\section{Related work}

\paragraph{Does post-training create new capabilities?} Comparisons at large sampling budgets find that RLVR mostly sharpens capabilities the base model already has~\citep{yue2025does}, with analogous conclusions for on-policy distillation~\citep{wang2026demystifying,li2026solving}; others report expansion under prolonged training~\citep{liu2025prorl}, relate apparent limits to the composition of primitive skills~\citep{wang2026newskills,zhang2025interplay}, or show that even spurious rewards improve Qwen2.5-Math~\citep{shao2025spurious}. We ask a different question: not whether the boundary moves in aggregate, but which problems move, and how predictable that is.

\paragraph{Per-problem outcomes of RLVR.} \citet{yang2026kernel} show, for public RLVR pairs, that problems with the same base success rate can have different post-RL success rates and that these differences repeat across independent generation halves of the same pair. We ask a complementary question: whether different runs on the same base model agree with each other on which problems improve, how much of that agreement any signal could predict, and whether it carries over to a different base model. \citet{chakma2026unlearnable} show that difficulty labels estimated from few rollouts are unstable, which motivates our many-sample design. \citet{chen2026unlearnability} identify training problems that remain unlearnable and observe that correct-rollout likelihoods are similarly distributed for learnable and unlearnable groups. \citet{zhou2026beyond} estimate a learnability prior for training problems from short probe training under one recipe. Our signals require no training, our outcomes are on held-out problems, and our ceiling is defined across recipes and groups.

\paragraph{Selecting and allocating problems.} Proposed signals include the student's likelihood of verified solutions~\citep{xu2026trek}, observed success counts~\citep{yuan2026diversity}, internal circuit activations~\citep{chen2026circuitlens} and prefix recoverability~\citep{li2026beyond}; teacher-guided curricula~\citep{zhu2026unlocking} and strategy-conditioned exploration~\citep{cui2026strategy} target problems that are currently unsolvable. A recent controlled comparison finds no rollout-selection method that reliably beats uniform sampling~\citep{liang2026dataflex}. Our measurements point to one reason: the closeness signals these methods build on capture only a minority of what post-training predictably changes. Reporting the noise ceiling alongside a signal's correlation would make the remaining headroom explicit.

\section{Conclusion}

Before post-training a model, can we tell which problems it will improve on? Across two base models and eighteen runs, the answer has three parts. What post-training fixes is shared across independent runs and specific to the base model. The closeness signals behind today's data selection and curricula capture only a minority of it. And the outcomes of a single existing checkpoint capture more than any a priori signal, alone or combined. The noise ceiling turns the remaining gap into a number. We hope it becomes a routine companion to per-problem signals, and that closing the gap becomes a goal for the next generation of selection methods.

\bibliographystyle{plainnat}
\bibliography{refs}

\appendix
\section{Experimental details}
\label{app:details}

\paragraph{Runs on Base B.} Qwen2.5-Math-1.5B-Oat-Zero~\citep{liu2025understanding}; One-Shot-RLVR-Qwen2.5-Math-1.5B-pi1, -7.5k-MATH and -1.2k-dsr-sub~\citep{wang2025oneshot} (one family); ExGRPO-Qwen2.5-Math-1.5B-Zero~\citep{zhan2025exgrpo}; ReLIFT-Qwen2.5-Math-1.5B-Zero~\citep{ma2025relift}; the step-1000 checkpoint of the spurious-rewards RLVR run~\citep{shao2025spurious}; and the public Hugging Face checkpoints relex-rlvr/RLVR-Qwen2.5-Math-1.5B, PRM-CoT/Qwen2.5-Math-1.5B-grpo-n5-step450 and yangzhch6/Zero-Solver-Qwen2.5-Math-1.5B-L. On the \FmNKept{} main-analysis problems their mean gains range from \FmGainMin{} to \FmGainMax{} percentage points.

\paragraph{Decontamination.} Problem statements are lower-cased, stripped of non-alphanumeric characters and truncated to 120 characters; a problem is removed if this key matches any problem in DAPO-17k or in the DeepScaleR, open-DeepScaleR, Open-RS, open-s1 and STILL-3 training sets. One of these sets alone contains 89\% of the Omni-MATH problems we considered. Training-set similarity is the maximum cosine similarity between TF-IDF vectors of character 4--5-grams of the problem and of any training problem.

\paragraph{Sampling and grading.} All models are served with vLLM using their own chat templates. Responses are graded by a rule-based verifier that extracts the final boxed answer and checks symbolic equivalence with the reference using \texttt{math-verify}. On Base A, \NOldNLowBaseSixFour{} of the rarely solved problems come from a second batch in which the base model was sampled 64 times; the disjoint-baseline analysis uses 64 further base samples per problem drawn with a different seed; 

\paragraph{OPD runs (Base A).} Student DeepSeek-R1-Distill-Qwen-1.5B, teacher JustRL-DeepSeek-1.5B, top-16 reverse KL on student rollouts, learning rate $10^{-6}$, batch size 64, maximum response length 7{,}168, 300 steps, using the public implementation of~\citet{fu2026oneshot}; the single-example run uses the example designated in that work. The OPD runs track their teacher closely ($\rho=\NOldLowCrossNoteacherROpdTeacher$ between their mean gain and JustRL's), which is why we group them with JustRL as one family.

\paragraph{Teacher-forcing.} Reference solutions are scored with the base model under its chat template, computing log-probabilities and ranks for every solution token; chunked computation matches a full-vocabulary computation exactly on a held-out check.

\section{Additional robustness results}
\label{app:robust}

\paragraph{Base B on all problems.} Without the format restriction, Base B has \FmNLow{} rarely solved problems; the noise ceiling for independent runs is $\CIndepCeiling$, the shared problem effect accounts for $\CVdIndepFracProblem$ of single-run variance, the combined a priori signals capture $\CSubsetExplainedBestModelPoint$ [$\CSubsetExplainedBestModelCiLo$, $\CSubsetExplainedBestModelCiHi$] of the explainable variance in gains, and a single checkpoint from a different group exceeds the best base-model signal by $\LtBaseTwoAllDiffBaseOnly$ [$\LtBaseTwoAllDiffBaseOnlyCiLo$, $\LtBaseTwoAllDiffBaseOnlyCiHi$] and the larger-model pass rate by $\LtBaseTwoAllDiffRefSeven$ [$\LtBaseTwoAllDiffRefSevenCiLo$, $\LtBaseTwoAllDiffRefSevenCiHi$].

\paragraph{Base A with all eight runs.} Including the JustRL lineage, the eight-run mean has reliability $\NOldLowRelEight$ and ceiling $\NOldLowCeiling$; the shared problem effect accounts for $\YVdLowAllFracProblem$ of single-run variance; rank-standardising each run before averaging gives a ceiling of $\XRelLowRankzCeiling$. Pooling more runs predicts a held-out run better: one, two, four and seven other runs reach $\XLoroLowByMOne$, $\XLoroLowByMTwo$, $\XLoroLowByMFour$ and $\XLoroLowMeanOthers$.

\begin{table}[h]
\centering
\small
\begin{tabular}{lccccc}
\toprule
& & \multicolumn{2}{c}{Outcome: gain} & \multicolumn{2}{c}{Outcome: post-training acc.} \\
\cmidrule(lr){3-4}\cmidrule(lr){5-6}
Base pass rate (Base A) & $N$ & Ceiling & Base pass rate $\rho$ & Ceiling & Base pass rate $\rho$ \\
\midrule
$(0, 25/256]$ & \XStrataZeroZeroOneN & $\XStrataZeroZeroOneLowCeiling$ & $\XStrataZeroZeroOneLowBaseAcc$ & $\XStrataZeroZeroOneLowaccCeiling$ & $\XStrataZeroZeroOneLowaccBaseAcc$ \\
$(25/256, 0.5]$ & \XStrataZeroOneZeroFiveN & $\XStrataZeroOneZeroFiveLowCeiling$ & $\XStrataZeroOneZeroFiveLowBaseAcc$ & $\XStrataZeroOneZeroFiveLowaccCeiling$ & $\XStrataZeroOneZeroFiveLowaccBaseAcc$ \\
$(0.5, 1)$ & \XStrataZeroFiveOneN & $\XStrataZeroFiveOneLowCeiling$ & $\XStrataZeroFiveOneLowBaseAcc$ & $\XStrataZeroFiveOneLowaccCeiling$ & $\XStrataZeroFiveOneLowaccBaseAcc$ \\
\bottomrule
\end{tabular}
\caption{Base A, all base pass-rate levels (eight runs). Base pass rate predicts post-training accuracy at every level but predicts gain only weakly, and negatively once little room is left.}
\label{tab:strata}
\end{table}

\paragraph{Per-run reliability (Base A).} Squared correlation of a run with the mean of the other seven, divided by the reliability of that mean: JustRL $\XRelLowPerRunJustrlRelSingle$, DeepScaleR $\XRelLowPerRunDeepscalerRelSingle$, ProRL $\XRelLowPerRunProrlRelSingle$, STILL-3 $\XRelLowPerRunStillThreeRelSingle$, Open-RS3 $\XRelLowPerRunOpenRsThreeRelSingle$, FastCuRL $\XRelLowPerRunFastcurlRelSingle$, OPD-full $\XRelLowPerRunOpdFullRelSingle$, OPD-1shot $\XRelLowPerRunOpdOneShotRelSingle$.

\paragraph{Reference-solution selection (Base A).} Per problem, statistics are taken from the reference solution with the highest worst-step log-probability. Averaging each statistic over all reference solutions gives worst-step $\XRefFairLowWorstStepAvg$, mean likelihood $\XRefFairLowMeanLikAvg$, top-16 count $\XRefFairLowRankOneSixAvg$ and length $\XRefFairLowLenAvg$ (magnitudes), and partialling out the number of reference solutions gives $\XRefFairLowWorstStepAvgPartialK$, $\XRefFairLowMeanLikAvgPartialK$, $\XRefFairLowRankOneSixAvgPartialK$ and $\XRefFairLowLenAvgPartialK$.

\paragraph{Semantic labels (Base A).} Adding the labels to the a priori signals (without training-set similarity, on the same problems and folds) moves the captured fraction from $\ZModelWithoutLabelsExplainedBestModelPoint$ [$\ZModelWithoutLabelsExplainedBestModelCiLo$, $\ZModelWithoutLabelsExplainedBestModelCiHi$] to $\ZModelWithLabelsExplainedBestModelPoint$ [$\ZModelWithLabelsExplainedBestModelCiLo$, $\ZModelWithLabelsExplainedBestModelCiHi$]; topic is also unrelated to the residual of gains after the two strongest signals ($p=\ZResidualTopicP$).

\end{document}